\pdfoutput=1

\documentclass[11pt]{article}

\usepackage[final]{acl}

\usepackage{times}
\usepackage{latexsym}

\usepackage[T1]{fontenc}
\usepackage[utf8]{inputenc}

\usepackage{microtype}

\usepackage{inconsolata}

\usepackage{graphicx}

\usepackage{multirow}
\usepackage{array}
\usepackage{pifont} % For star symbols
\usepackage{booktabs}
\usepackage{amsmath}
\usepackage{placeins}
\usepackage{stfloats}
\usepackage{float}
\usepackage{amsfonts}

\title{Designing and Contextualising Probes for African Languages}

\author{Wisdom Aduah \\
  	African Institute for Mathematical Sciences \\
    South Africa\\
  \texttt{wizzy@aims.ac.za} \\\And
  Francois Meyer \\
  Department of Computer Science \\
  University of Cape Town \\
  \texttt{francois.meyer@uct.ac.za} \\}

\usepackage{times}
\usepackage{latexsym}

\usepackage{booktabs}
\usepackage{siunitx}
\usepackage{amsmath,amssymb,amsthm}
\usepackage{multirow}

\usepackage{array}
\newcommand{\PreserveBackslash}[1]{\let\temp=\\#1\let\\=\temp}
\newcolumntype{C}[1]{>{\PreserveBackslash\centering}p{#1}}
\newcolumntype{R}[1]{>{\PreserveBackslash\raggedleft}p{#1}}
\newcolumntype{L}[1]{>{\PreserveBackslash\raggedright}p{#1}}
\begin{document}
\maketitle
\begin{abstract}

Pretrained language models (PLMs) for African languages are continually improving, but the reasons behind these advances remain unclear. This paper presents the first systematic investigation into probing PLMs for linguistic knowledge about African languages. We train layer-wise probes for six typologically diverse African languages to ana\-lyse how linguistic features are distributed. We also design control tasks, a way to interpret probe performance, for the MasakhaPOS dataset. We find PLMs adapted for African languages to encode more linguistic information about target languages than massively multilingual PLMs. Our results reaffirm previous findings that token-level syntactic information concentrates in middle-to-last layers, while sentence-level semantic information is distributed across all layers. Through control tasks and probing baselines, we confirm that performance reflects the internal know\-ledge of PLMs rather than probe memorisation. Our study applies established interpretability techniques to African-language PLMs. In doing so, we highlight the internal mechanisms underlying the success of strategies like active learning and multilingual adaptation.
\end{abstract}

\section{Introduction}

\begin{figure}[t]
    \centering
    
        \includegraphics[width=0.95\linewidth, trim={0.6cm 0.2cm 0.7cm 1cm},clip]{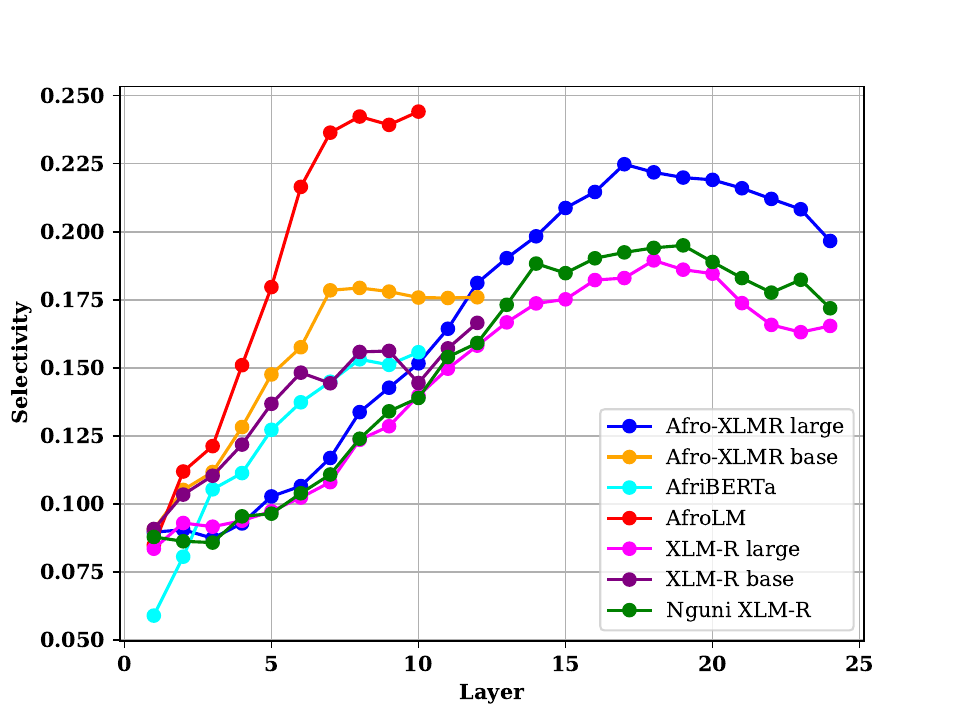}
    
    \caption{POS probe performance (selectivity), averaged over 6 African languages.}
    \label{fig:pos_control-label}
\end{figure}
\begin{figure}[h!]
    \centering
    
        \includegraphics[width=0.9\linewidth, trim={1cm 0.2cm 1cm 1.3cm},clip]{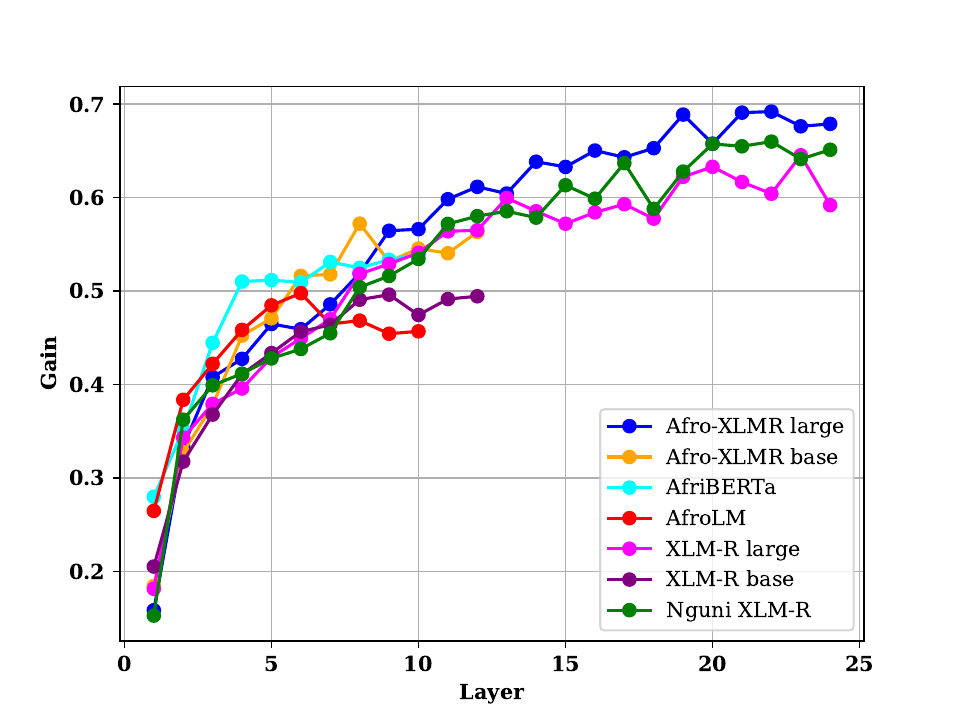}
    
    \caption{NER probe gains (over random baselines) across layers, averaged over 6 African languages.}
    \label{fig:ner_control-label}
\end{figure}

The past few years have seen the proliferation of pretrained language models (PLMs) across various domains including education, healthcare, and finance \citep{hadi2024large}. %While demonstrating impressive capabilities in natural language understanding and generation, 
The blackbox nature of these models, paired with their increasing size and complexity, has prompted the growing subfield of NLP interpretability \citep{luo2024understandingutilizationsurveyexplainability}. These methods aim for insights into the internal mechanisms underlying the success and failures of PLMs.
One of the earliest interpretability methods to gain traction in NLP was probing \citep{alain2018understandingintermediatelayersusing}, which trains a classifier on intermediate PLM representations. Probing measures to what extent specific linguistic features, such as part-of-speech (POS) categories or semantic concepts, are encoded in hidden layers. 
%This reveals in which layers a model encodes different aspects of linguistic knowledge. 
%Such insights enable a better understanding of internal model decision-making, helping to identify failure modes and highlighting opportunities for improvement.

\begin{table*}[t] \small
    \centering
    \begin{tabular}{lccC{2.9em}C{2.9em}C{2.9em}C{3.3em}C{3em}C{2.9em}}
        \toprule
       \textbf{Model} & \textbf{Layers} & \textbf{Params} & \textbf{Swahili} & \textbf{Igbo} & \textbf{Hausa} & \textbf{Luganda} & \textbf{isiXhosa} & \textbf{Naija} \\ 
         % \textbf{Family} &  &  & Bantu & Volta-Niger & Chadic & Bantu & Bantu & Creole \\ 
         %  \textbf{Region} &  &  & East & West & West & East & South & West \\ 
        %\midrule
        % \textbf{Model} & \textbf{Layers} & \textbf{Params} & &  & &  &  & \\        
        \midrule
       XLM-R-base  \citep{conneau-etal-2020-unsupervised} & 12&270M   & \ding{72}\ding{72} & 
        \ding{73}\ding{73} & 
        \ding{72}\ding{72} & 
        \ding{73}\ding{73} & 
        \ding{72}\ding{72} & 
        \ding{73}\ding{73} \\
       XLM-R-large \citep{conneau-etal-2020-unsupervised} & 24&550M  & \ding{72}\ding{72} & 
        \ding{73}\ding{73} & 
        \ding{72}\ding{72} & 
        \ding{73}\ding{73} & 
        \ding{72}\ding{72} & 
        \ding{73}\ding{73} \\
       AfroXLMR-base \citep{alabi-etal-2022-adapting}  & 12&270M  & \ding{72}\ding{72} & 
        \ding{72}\ding{72} & 
        \ding{72}\ding{72} & 
        \ding{73}\ding{73} & 
        \ding{72}\ding{72} & 
        \ding{72}\ding{72} \\        
        AfroXLMR-large \citep{alabi-etal-2022-adapting}  &  24&550M& \ding{72}\ding{72} & 
        \ding{72}\ding{72} & 
        \ding{72}\ding{72} & 
        \ding{73}\ding{73} & 
        \ding{72}\ding{72} & 
        \ding{72}\ding{72} \\        
          Nguni-XLMR \citep{meyer-etal-2024-nglueni} & 24&550M   & \ding{72}\ding{73} & 
        \ding{72}\ding{73} & 
        \ding{72}\ding{73} & 
        \ding{73}\ding{73} & 
        \ding{72}\ding{72} & 
        \ding{73}\ding{73} \\
        AfriBERTa  \citep{ogueji-etal-2021-small}  &  10&126M & \ding{72}\ding{72} & 
        \ding{72}\ding{72} & 
        \ding{72}\ding{72} & 
        \ding{73}\ding{73} &
        \ding{73}\ding{73} & 
        \ding{72}\ding{72} \\
        AfroLM \citep{dossou-etal-2022-afrolm} & 10&264M   & \ding{72}\ding{72} & 
        \ding{72}\ding{72} & 
        \ding{72}\ding{72} &
        \ding{72}\ding{72} & 
        \ding{72}\ding{72} & 
        \ding{72}\ding{72} \\

        \bottomrule
    \end{tabular}
    \caption{Language coverage of PLMs. \ding{73}\ding{73} indicates no data from the language was included in pretraining or adaptation. \ding{72}\ding{73} shows the language was included in the base model but not in the adapted model. \ding{72}\ding{72} shows the model was either pretrained or adapted for the language.}
    \label{tab:lang_coverage}
\end{table*}

Probing provides insights into the internal mechanisms of PLMs by revealing how models acquire, store, and leverage linguistic information in hidden layers. This allows NLP practitioners to better understand the mechanisms by which PLMs succeed in certain tasks, and can also point to the underlying reasons for failing in others. Besides contributing to a greater, linguistically grounded understanding of PLM computations, probing also has the potential to contribute to performance gains by guiding the finetuning process for downstream tasks. For example, knowing which layers encode specific properties can inform which layers should be targetted for finetuning, optimising both performance and efficiency \citep{53741be929584774a704bbccf7ae71b9}. 

Probing is an established tool in NLP interpretability, having been extensively applied and studied across different settings. One area where it has yet to be applied is the growing body of work on PLMs for African languages. Most African languages are under-represented in the pretraining data of multilingual PLMs, which limits their performance. 
%This is a pressing issue, as the lack of reliable NLP tools for African languages restricts home language access to technology for most of the African continent. 
Efforts to address this gap have led to the development of PLMs targeting African languages, such as AfriBERTa \citep{ogueji-etal-2021-small}, AfroLM \citep{dossou-etal-2022-afrolm}, and AfroXLMR \cite{alabi-etal-2022-adapting}. These models leverage strategies such as cross-lingual transfer \citep{conneau-etal-2020-unsupervised}, active learning \citep{dossou-etal-2022-afrolm}, and multilingual adaptive fine-tuning (MAFT) \citep{alabi-etal-2022-adapting} to improve performance for low-resource languages.

Despite this progress, there is limited understanding of how these PLMs encode African languages internally, which is where probing holds promise. Most probing research targets higher-resourced languages such as English, French, and Russian \citep{arps-etal-2024-multilingual, 
%zhang-bowman-2018-language, 
%hewitt2019designing, 
53741be929584774a704bbccf7ae71b9, conneau-etal-2018-cram, hou-etal-2024-transformers}. Previous works have explored some low-resource languages, such as Tagalog, Hindi and Tamil \citep{Arora2022ProbingPL, li2024exploringmultilingualprobinglarge}, but to the best of our knowledge, there has been no research targeting African languages.

% While there has been s
% Probing is an established technique in the NLP interpretability toolbox.  

In this paper, we conduct the first systematic probing study for PLMs focussed on African languages. 
We design probes for POS tagging, named entity recognition (NER), and news topic classification (NTC), using the MasakhaPOS \citep{dione-etal-2023-masakhapos}, MasakhaNER \citep{adelani-etal-2022-masakhaner}, and MasakhaNEWS \citep{Adelani2023MasakhaNEWSNT} datasets respectively. 
We train probes on seven masked PLMs (listed in \autoref{tab:lang_coverage}), representing different approaches to developing PLMs for low-resource languages. We evaluate how syntactic and semantic information is encoded for six African languages, which cover different language families and varying levels of data availability, as shown in \autoref{tab:lang_coverage}.

To interpret probe accuracies, one has to isolate the contribution of model-encoded knowledge, as opposed to the probe itself learning the task. To enable such probe interpretability for African languages, we design a control task \citep{hewitt2019designing} 
for MasakhaPOS. Control tasks are synthetic tasks that measure to what extent probes can learn a task without model-encoded knowledge. Our control task enables researchers to contextualise probing results for MasakhaPOS.
%$This provides researchers with a way to contextualise probing results for MasakhaPOS. 
%To contextualise probing results, it is common to train probes on randomly initialised representations as a baseline. 
% We To the same end \citet{hewitt2019designing} propose control tasks, the creation of artificial tasks in which input tokens are randomly assigned to labels. Control tasks are independent of linguistic information, so probe accuracies reflect to what extent probes can succeed without access to model-encoded linguistic features. 
% To enable the same probe interpretability for African languages, we design a control task for MasakhaPOS, our POS tagging probing task. This provides researchers with a way to contextualise probing results for MasakhaPOS. 

Our main findings can be summarised as follows:
\begin{enumerate}
 \item Word-level linguistic knowledge (POS, NER) concentrates in middle layers, while sentence-level information (NTC) is spread out.
     \item The inclusion of target languages in pretraining or multilingual adaptation improves probe performance across all tasks.
     \item Cross-lingual transfer improves probe performance for languages not in pretraining, but less so for low-resource language families.
\end{enumerate}

\section{Background}
\label{sec:background}

\paragraph{PLMs for African Languages}
Multilingual modelling has been leveraged in different ways to build PLMs for African languages. The massively multilingual XLM-R \citep{conneau-etal-2020-unsupervised} is trained on 100 languages, of which only 8 are African languages. 
%However, XLM-R covers only 8 African languages (0.2\% of its pretraining data). 
AfroXLMR \citep{alabi-etal-2022-adapting} improves performance by adapting XLM-R for 17 African languages, while 
Nguni-XLMR \citep{meyer-etal-2024-nglueni} adapts XLM-R for the four Nguni languages (isiXhosa, isiZulu, isiNdebele, and Siswati). 
AfriBERTa \citep{ogueji-etal-2021-small} is a smaller model pretrained from scratch on 11 African languages on less than 1GB data. 
AfroLM \citep{dossou-etal-2022-afrolm} is also trained from scratch on 23 African languages, using self-active learning (the model learns to identify beneficial training samples). 

% is a cross-lingual masked language model based on RoBERTa \citep{conneau-etal-2020-unsupervised}, trained on approximately 2.5 TB of the Common Crawl corpus covering 100 languages. However, only 8 African languages are included, making up just 0.2\% of the dataset. 

\paragraph{Contextualising Probe Performance}
Probes are not direct measures of model-encoded knowledge, since the probe itself can contribute to performance by learning the task. 
% One of the challenges in probing is interpreting the results. For instance, if a probe achieves an accuracy of 80\%, it is not clear whether this should be considered high or low without a suitable baseline, as the probe could have simply memorized the task.
Probing studies use baselines, such as majority class prediction \citep{belinkov2017neural, conneau-etal-2018-cram} or probes trained on random representations \citep{zhang2018language, conneau-etal-2018-cram, chrupala2020analyzing, tenney2019you}, to contextualize performance. 

However, even random baselines may encode information that a sophisticated classifier could exploit. As an alternative, \citet{hewitt2019designing} propose control tasks: pairing word types with random labels to neutralise the linguistic information required for the original task. They define selectivity as the difference between original task and control task accuracy. Selectivity captures the contribution of linguistic knowledge to probe performance, so it is a more reliable measure of model knowledge than raw probe accuracies. To enable probe contextualisation for African languages, we design a control task for the MasakhaPOS dataset.

% Alternatively, \citet{pimentel2020information} suggest measuring information gain through a control function applied to internal representations, providing a more generalizable framework beyond word-level tasks.
% \citet{ravichander2020probing} introduce control datasets for contextualizing probing results, while other studies \citep{alain2018understandingintermediatelayersusing, hewitt2019designing} advocate restricting the probe’s complexity and training data size to improve interpretability.

\section{Probing Framework}

% \begin{figure*}[t]
%     \centering
%     \includegraphics[width=0.9\linewidth]{Probing framework.drawio-6.pdf}
%     \caption{Probing task illustration. \(h_1, h_2, \text{ and } h_3 \text{ are the hidden state representation in an arbitrary layer.}\) 'worked' is tokenized into '\_work' and 'ed', but the representation for '\_work' (first subword pooling) is fed to the classifier.}
%     \label{fig:probe_illustration}
% \end{figure*}

\subsection{Probe Design}

Some works advocate for linear probes \citep{alain2018understandingintermediatelayersusing, hewitt2019designing}, arguing that they are less prone to memorisation, while others argue that some linguistic features might not be linearly separable in the representation space \citep{conneau-etal-2018-cram, pimentel2020information}. 

For our experiments, we select a probe to strike a balance between complexity and simplicity. 
%It is designed to be not overly complex, reducing the risk of memorization, yet not too simple, allowing it to capture linguistically relevant features that are not linearly separable. 
Our probe classifier is a multilayer perceptron (MLP) with a single hidden layer of 50 neurons, which we formally define as
\begin{align}
\mathbf{y} = f(\mathbf{W_2} \, \sigma(\mathbf{W_1} \mathbf{x} + \mathbf{b_1}) + \mathbf{b_2}),
\end{align} where \( \mathbf{x} \in \mathbb{R}^n \) is the input representation, \( \mathbf{W_1} \in \mathbb{R}^{m \times n} \) is the weight matrix for the hidden layer, \( \mathbf{W_2} \in \mathbb{R}^{k \times m} \) is the weight matrix for the output layer, \( \mathbf{b_1} \in \mathbb{R}^m \) and \( \mathbf{b_2} \in \mathbb{R}^k \) are bias vectors, \( \sigma(\cdot) \) is the ReLu activation function, and \( f(\cdot) \) is a softmax function for label prediction. %\(m\text{, } n\text{, and }k\) are the number of units in the hidden layer, the embedding dimensionality, and number of units in the output layer respectively.

For POS tagging and NER, we define a word-level task as a function \(f\) that maps an input sequence \(X\) to an output sequence \(Y\). That is \(f:X\longrightarrow Y\),
where \(X\) is a sequence of contextualized hidden representations (embeddings) of the input text, and \(Y\) is the sequence of output labels corresponding to the words encoded by \(X\). %For example, given the input sequence [He, worked, alone], the pretrained model generates contextualized embeddings for each subword in each layer. The probe then predicts the label for each subword embedding for a specific layer as illustrated in Figure \ref{fig:probe_illustration}. 
Given that some words are tokenized into multiple subwords, we use the first subword in each word to represent the word in the classifier. %as input for the classifier to align the words with their hidden representations. 

For news topic classification (NTC), we define a sentence-level task similarly. Instead of passing word embeddings to the probe classifier, we pass the embedding of the special token for sequence classification  (e.g. \texttt{<s>} for XLM-R).
% special classification token are passed as input to the classifier, since they represent the entire input sequence. 
We truncate inputs consisting of more tokens than the maximum sequence length of PLMs.
%For input texts longer than the maximum input token length for the model, the texts were truncated to fit the desired input size.

%\subsection{Subword pooling strategy}
\subsection{MasakhaPOS Control Task}
\label{subsec:controltask}

As discussed in \autoref{sec:background}, control tasks \citep{hewitt2019designing} can be used to contextualise probe results. A probe could achieve a high raw accuracy by learning to map word types to labels, without relying on linguistic knowledge. For example, a probe classifier could learn to map the embedding of ``walk'' to the POS tag ``verb'', by learning the mapping between word type and label (instead of the mapping between syntactic role and label). 
% \label{sec:control_tasks}
% Since the probe functions as a classifier, it is prone to memorizing dataset patterns unrelated to the linguistic property being studied. A more meaningful evaluation of its performance should account for this memorization effect rather than relying solely on raw accuracy. Control tasks \citep{hewitt2019designing} help address this by requiring the model to solve a task purely through memorization of input-output pairs, providing a baseline for comparison. 
\citet{hewitt2019designing} propose \emph{selectivity} as an alternative to raw accuracy. Selectivity is defined as the difference between linguistic task accuracy and the control accuracy. As a measure, it isolates the contribution of model-encoded linguistic knowledge to probe performance. 

A control task is designed in two steps: 
\begin{enumerate}
    \item Define the random control behavior for each word type i.e. assign a label \emph{randomly} to each word in the vocabulary.
    \item Deterministically label the original task corpus based on control behaviour i.e. annotate each word with its assigned random label.
\end{enumerate}

To define a control task for MasakhaPOS, we randomly map each unique word in the dataset to a random POS tag. Next, we use this synthetic mapping to re-annotate the train / validation / test set. As per \citet{hewitt2019designing}, when creating the random mapping (the control behaviour) we sample POS tags according to their empirical distribution in the original MasakhaPOS dataset. Control tasks are designed to have both structure and randomness. Our MasakhaPOS contains structure in that a word type is always mapped to the same tag, but the assignment is random in that it is independent of the linguistic role of words.

\subsection{NTC and NER Probe Baselines}
\label{sec:random_baselines}

Control tasks define word-level control behaviour, so they are not applicable to sentence-level tasks. To interpret NTC probes, we compare the performance of probes trained on PLMs to those trained on random contextual representations. 
%Therefore, we included probes trained on randomly initialized representations of the models, as random baselines for these tasks. 
Following \citet{hewitt-manning-2019-structural}, we use an untrained bi-LSTM, and mean-pooling word-level outputs to produce a single sentence representation. Probes trained on these outputs can leverage contextual information, but no linguistic knowledge.

NER is a word-level task, so controls tasks could plausibly be designed for MasakhaNER. However, the procedure is complicated by the fact that NER is actually a \emph{span-level} task (named entities can span multiple words). It is not obvious how to extend the control behaviour design of \citet{hewitt2019designing} to multi-word spans. To contextualise NER results, we randomly re-initialise the architectures of our seven probed PLMs to serve as random baselines \citep{zhang2018language, conneau-etal-2018-cram}. 
We estimate model-encoded knowledge by subtracting the F1 score of a random model from the F1 score of the corresponding PLM. For each layer, we refer to this as the probe \emph{gain} over the random baseline.

% which we defined as the F1 score of the PLM 

% \begin{multline*}
%     \textit{gain} = \textit{f1-score on original model} - \\ \textit{f1-score on randomly initialized model}
% \end{multline*}

\section{Experimental Setup}

\subsection{Data}

Both MasakhaPOS \citep{dione-etal-2023-masakhapos} and MasakhaNER \citep{adelani-etal-2022-masakhaner} cover 20 African languages. MasakhaNEWS \citep{Adelani2023MasakhaNEWSNT} covers 16 African languages and contains news articles annotated with one of seven topic labels (business, entertainment, health, politics, religion, sports, technology). 

\subsection{Languages}

The six language in our study (Swahili, Igbo, Hausa, Luganda, isiXhosa, and Naija Pidgin) are included in all three Masakhane\footnote{\url{https://www.masakhane.io/}} datasets. We chose these languages specifically to cover several language families, a broad range of data availability, and varying levels of representation in existing PLMs (as shown in \autoref{tab:lang_coverage}). As shown in \autoref{tab:langs}, the languages cover four language families across East, West, and Southern Africa.

\begin{table}[t] \small
    \centering
    \begin{tabular}{lccc}
        \toprule
       \textbf{Language} &  \textbf{Family} & \textbf{Region} & \textbf{mC4 tokens}\\
        \midrule
        Swahili      & Bantu  & East & 1B\\
        Igbo         & Volta-Niger  & West & 90m \\
        Hausa        & Chadic & West & 200m \\
        Luganda      & Bantu  & East &  0\\
        isiXhosa     & Bantu  & Southern & 60m\\
        Naija Pidgin & Creole       & West & 0\\
        \bottomrule
    \end{tabular}
    \caption{The languages used in our study. The number is tokens in the mC4 corpus \citep{xue-etal-2021-mt5} serves to give an indication of broader data availability.}
    \label{tab:langs}
\end{table}

\subsection{PLMs}

The seven PLMs selected for our study represent established approaches to developing PLMs for African languages. XLM-R-base and XLM-R-large \citep{conneau-etal-2020-unsupervised} employ massively multilingual pretraining, while AfroXLMR-base, AfroXLMR-large \citep{alabi-etal-2022-adapting}, and Nguni-XLMR-large \citep{meyer-etal-2024-nglueni} adapt XLM-R to a more limited set of African languages. Afro\-XLMR takes a broader adaptation approach than Nguni-XLMR, which focusses only on the four Nguni languages, a group of related languages which includes isiXhosa. AfriBERTa \citep{ogueji-etal-2021-small} represents memory-efficient pretraining -- it is our smallest model both in terms of parameters and training data size.  AfroLM \citep{dossou-etal-2022-afrolm} represents sample-efficient pretraining, since its self-active learning maximizes available data by identifying beneficial training samples.

\section{Results}
\begin{figure*}[!h]
    \centering
    \makebox[\textwidth][c]{%
        \includegraphics[width=1.1\textwidth, trim={1.5cm 1.5cm 1.5cm 2cm},clip]{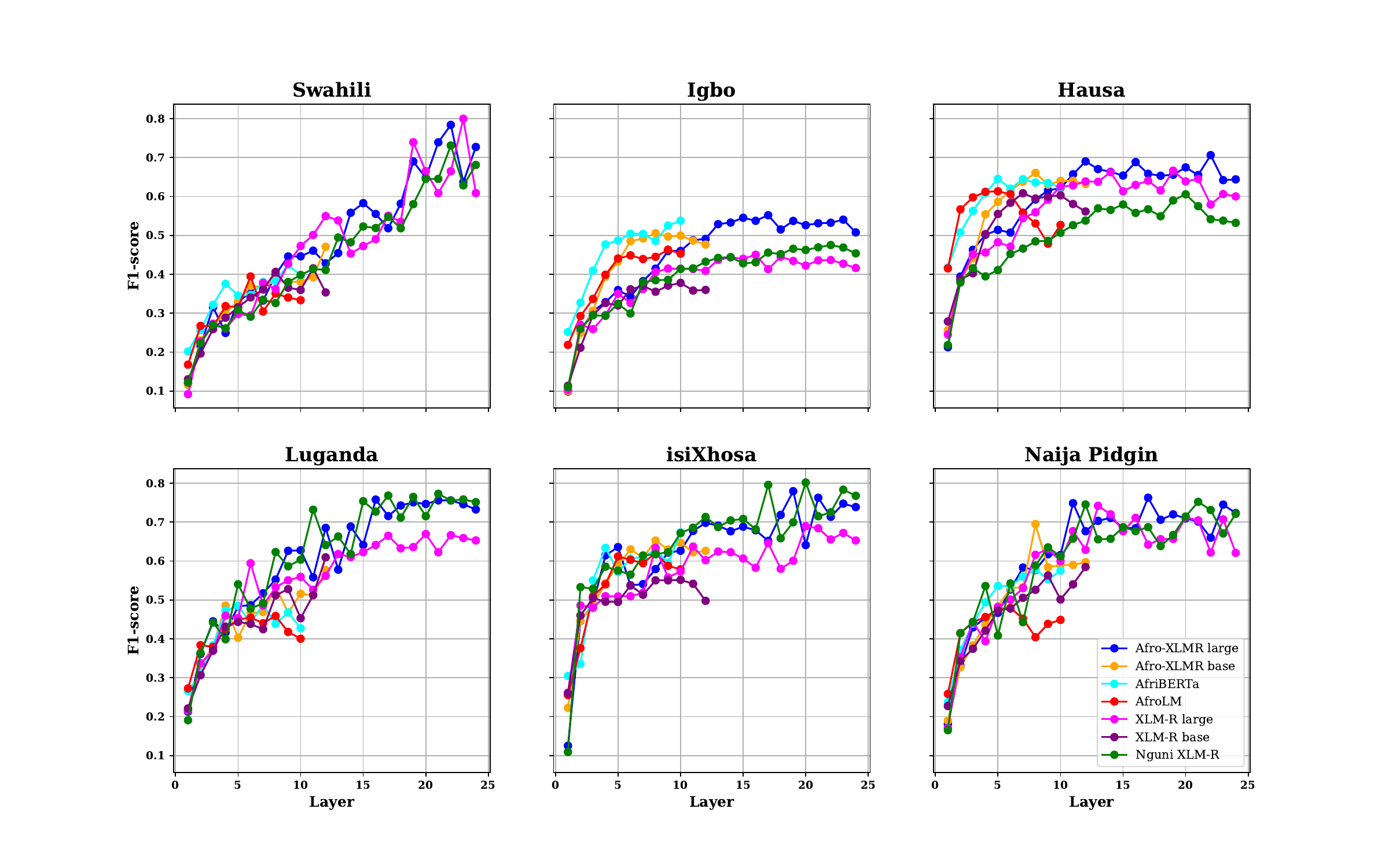}
    }
    \caption{Probe performance gains for NER tagging (F1 improvements over randomly re-initialised PLM architectures), across all layers and 6 African languages.}
    \label{fig:ner_gain-label}
\end{figure*}

We plot probing results for POS, NER, and NTC respectively in \autoref{fig:pos_control-label}, \autoref{fig:ner_gain-label}, and \autoref{fig:ntc_raw-label}. We report and compare best-layer results for each language, model, and task in \autoref{tab:aggregated_results}. %The full proving results are included in \autoref{sec:supplementary_results}.

\begin{figure*}[t]
    \centering
    \makebox[\textwidth][c]{%
        \includegraphics[width=1.1\textwidth, trim={1.5cm 1.5cm 1.5cm 2cm},clip]{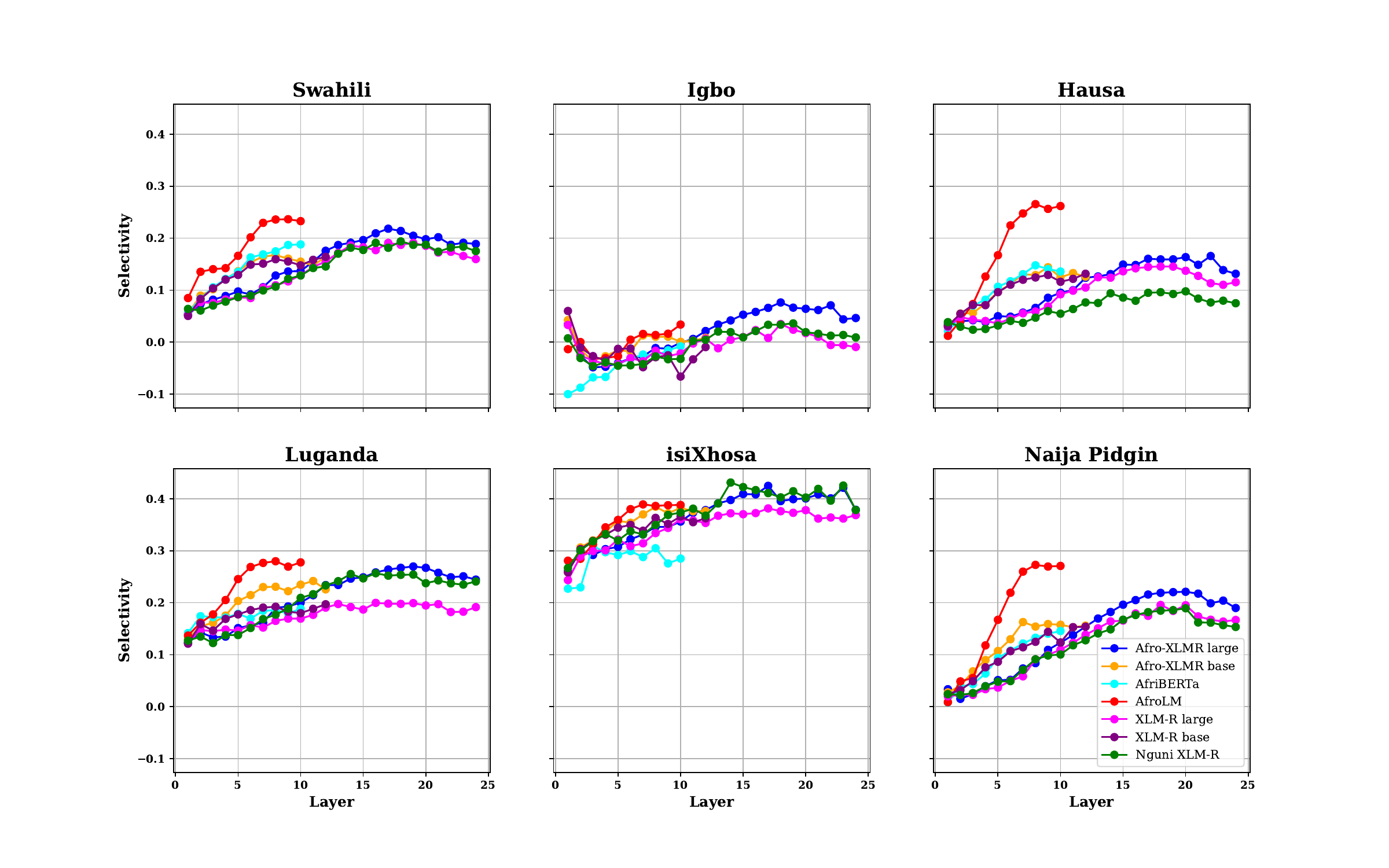}
    }
    \caption{Probe selectivity for POS tagging (the difference between MasakhaPOS accuracy and control task accuracy), across all layers and 6 African languages.}
    \label{fig:pos_control-label}
\end{figure*}

\subsection{POS Tagging}

We evaluate our POS probes based on selectivity, which is computed using the MasakhaPOS control task described in \autoref{subsec:controltask}. As shown in \autoref{fig:pos_control-label}, the PLMs exhibit positive selectivity across all layers for all languages, except in the case of Igbo. This aligns with previously reported PLM results for MasakhaPOS \citep{dione-etal-2023-masakhapos}, where POS tagging accuracies for Igbo were lower than all other languages. Igbo is from the Volta-Niger family, which is under-represented in the datasets of all seven models (as shown in \autoref{tab:language_distribution_family} in the appendix). This limits the benefit of cross-lingual transfer for Igbo.

For all other languages, POS selectivity is consistently positive, indicating that syntactic roles are reliably being encoded in the hidden representations of the PLMs. A clear pattern emerges in the distribution of POS information across layers. Probe selectivity is low in early layers, increases steadily in middle layers, peaks and plateaus in deeper layers, and finally decreases slightly in final layers. This pattern aligns with existing literature showing that middle-to-last-layers encode syntactic features more effectively \citep{rogers-etal-2020-primer}.

AfroLM stands out as encoding a high amount of POS information, achieving the highest selectivity overall on four of the six languages. While the exact reason for this is unknown, it is possible that self-active learning is used to select training examples that improve the model's syntactic knowledge during pretraining. Among the deeper models, AfroXLMR-large exhibits the greatest internal synactic knowledge overall, even achieving reasonable selectivity scores for Igbo in deeper layers. 
The difference in selectivity between AfroXLMR and XLM-R highlights the ability of multilingual adaptation to encode linguistic knowledge about specific languages. Similarly, 
Nguni-XLMR, exhibits the best probe performance for isiXhosa, one of its four target languages.

We include the raw probe accuracies for POS tagging in \autoref{fig:raw_pos_accuracies} in the appendix. The accuracies are quite high, comparing favourably to previously reported PLM performance for MasakhaPOS \citep{dione-etal-2023-masakhapos}.

% \newpage
\subsection{NER}

To contextualise our NER probes, we compute the per-layer difference between the F1 scores of probes trained on PLMs and their re-initialised counterparts (described in \autoref{sec:random_baselines}). As shown in \autoref{fig:ner_gain-label}, probes trained on PLMs consistently exhibit performance gains over random baselines across all layers and languages. The general trends observed for NER probes are similar to those of POS probes. AfroXLMR achieves the highest probe gains across different languages, while Nguni-XLMR does particularly well for isiXhosa. As in POS tagging, probe performance peaks in middle-to-later layers. 

We also observe evidence of cross-lingual knowledge representation. Luganda is not included in the pretraining data of either AfroXLMR or Nguni-XLMR but both exhibit high probe performance gains for Luganda than AfroLM, which is pretrained on Luganda. Luganda is of the Bantu language family, which is better represented than other families in the pretraining data of our PLMs (as shown in \autoref{tab:language_distribution_family} in the appendix). This shows that the PLMs are encoding linguistic similarities between different languages. This cross-lingual representation learning is the mechanism behind the impressive zero-shot performance of PLMs previously reported on MasakhaNER \citep{adelani-etal-2022-masakhaner}.

\subsection{News Topic Classification (NTC)} 

\begin{figure*}[t]
    \centering
    \makebox[\textwidth][c]{%
        \includegraphics[width=1.1\textwidth, trim={1.5cm 1.5cm 1.5cm 2cm},clip]{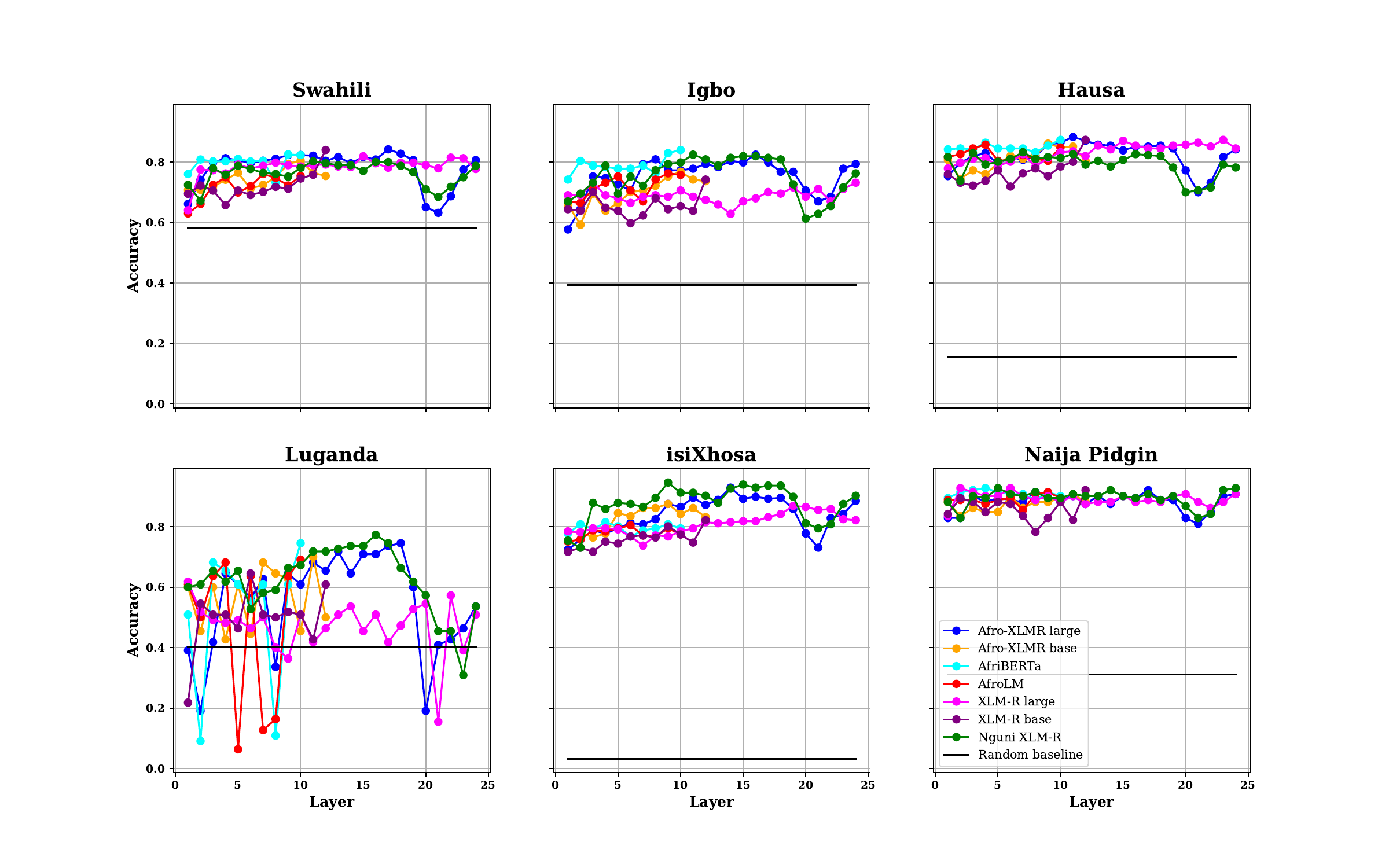}
    }
    \caption{Probe accuracy for news topic classification (visualised in comparison to a random contextual baseline) across all layers and 6 African languages.}
    \label{fig:ntc_raw-label}
\end{figure*}

To contextualise our NTC probes, we compare the classification accuracies of probes trained on PLMs and probes trained on random, contextual baselines (described in \autoref{sec:random_baselines}). \autoref{fig:ntc_raw-label} plots probe accuracies alongside random baseline performance. As for POS and NER, multilingual adaptation enhances sentence-level representations for target languages. Beyond this, NTC probing results diverge from the trends reported for POS and NER. 

Probe accuracy remains relatively consistent across layers, which aligns with previous work showing that sentence-level semantic information is spread across layers \citep{tenney-etal-2019-bert}. The one exception to this is Luganda, which exhibits high variance across layers and is the only language for which some PLM layers fall below random probe performance. We are unable to explain this behaviour. It is possible that the data scarcity of Luganda (see \autoref{tab:langs}) is a contributing factor and that, unlike for syntactic knowledge, cross-lingual semantic knowledge does not transfer as effectively.

\subsection{Analysing Trends Across Tasks}

\autoref{tab:aggregated_results} lists results for the top-performing layer for each PLM, across all languages and tasks. For each PLM and language, it also shows to what extent the language is represented by the model: (1) not included at all, (2) included in pretraining but not adaptation, or (3) included in pretraining or adaptation. The table reveals trends that hold across all three tasks. 

Multilingual adaptation is known to be a reliable method to improve downstream performance for low-resource languages. Our results show that this is being achieved, in part, by enhancing internal representations of target languages. 
AfroXLMR-large and Nguni-XLMR-large have the most instances of top-performing layers (six cases each). AfroXLMR-large performs best across Swahili, Igbo, and Hausa, all three of which belong to different language families. The multilingual adaptation of AfroXLMR is highly effective at enhancing linguistic feature encoding across typologically diverse languages. Nguni-XLMR-large performs best for isi\-Xhosa and Luganda (which is also of the Bantu language family). The more focussed, linguistically oriented adaptation of Nguni-XLMR effectively enhances linguistic knowledge for a more limited set of related languages. 

A clear pattern in \autoref{tab:aggregated_results} is the fact that all top-performing layers (except two) occur in PLMs that represent probed languages in their final pretraining stage (either during adaptation or in pretraining without subsequent adaptation). Best-layer performances (\textbf{boldface} in the table) almost always co-occur with maximal language representation (\ding{72}\ding{72}). The only exception to this is Luganda, for which Nguni-XLMR-large achieves two best-layer results. While we have previously discussed evidence of zero-shot cross-lingual representation learning, it is clear that including languages in pretraining is essential for encoding language-specific syntactic and semantic knowledge.

\begin{table*}[h]
    \centering \small
    \begin{tabular}{L{3cm}|l|C{3em}C{3em}C{3em}C{3.5em}C{3em}C{3em}}
        \toprule
        
        & & \textbf{Swahili} & \textbf{Igbo} & \textbf{Hausa} & \textbf{Luganda} & \textbf{isiXhosa} & \textbf{Naija} \\ 
        \midrule
          \multirow{4}{*}{\textbf{XLM-R-base}} 
           & Language coverage & \ding{72}\ding{72} & 
        \ding{73}\ding{73} & 
        \ding{72}\ding{72} & 
        \ding{73}\ding{73} & 
        \ding{72}\ding{72} & 
        \ding{73}\ding{73} \\ 
        & POS selectivity & 16.39 & 5.98 & 13.15 & 19.23 & 36.32 & 15.39 \\ 
       & NER gain & 41.54 & 37.79 & 60.84 & 60.95 & 55.16 & 58.46 \\ 
       & NTC accuracy & 84.03 & 74.23 & 87.38 & 64.55 & 82.15 & 92.11 \\ 
         \midrule
        \multirow{4}{*}{\textbf{XLM-R-large}} 
        & Language coverage & \ding{72}\ding{72} & 
        \ding{73}\ding{73} & 
        \ding{72}\ding{72} & 
        \ding{73}\ding{73} & 
        \ding{72}\ding{72} & 
        \ding{73}\ding{73} \\
       & POS selectivity & 19.09 & 3.29 & 14.54 & 19.97 & 38.15 & 19.56 \\ 
            & NER gain& \textbf{79.98} & 45.03 & 66.63 & 66.95 & 68.97 & 74.20 \\ 
          & NTC accuracy & 81.93 & 73.20 & 87.38 & 61.82 & 86.87 & 92.76 \\ 
         \midrule
      \multirow{4}{*}{\textbf{AfroXLMR-base}} 
      & Language coverage & \ding{72}\ding{72} & 
        \ding{72}\ding{72} & 
        \ding{72}\ding{72} & 
        \ding{73}\ding{73} & 
        \ding{72}\ding{72} & 
        \ding{72}\ding{72}\\ 
               & POS selectivity  & 16.73 & 4.22 & 14.40 & 24.18 & 38.48 & 16.28 \\ 
        & NER gain& 47.07 & 50.55 & 66.06 & 57.7 & 65.27 & 69.52 \\ 
         & NTC accuracy & 80.46 & 76.80 & 86.12 & 70.00 & 87.54 & 90.79 \\ 
         \midrule
        \multirow{4}{*}{\textbf{AfroXLMR-large}} 
        & Language coverage & \ding{72}\ding{72} & 
        \ding{72}\ding{72} & 
        \ding{72}\ding{72} & 
        \ding{73}\ding{73} & 
        \ding{72}\ding{72} & 
        \ding{72}\ding{72}\\
               & POS selectivity  & 21.82 & \textbf{7.62} & 16.56 & 26.97 & 42.49 & 22.03 \\ 
         & NER gain & 78.41 & \textbf{55.20} & \textbf{70.63} & 75.79 & 77.96 & \textbf{76.27} \\
         & NTC accuracy & \textbf{84.24} & 82.47 & \textbf{88.33} & 74.55 & 92.93 & 90.79 \\ 
        \midrule
        
        \multirow{4}{*}{\textbf{Nguni-XLMR-large}} 
        & Language coverage & \ding{72}\ding{73} & 
        \ding{72}\ding{73} & 
        \ding{72}\ding{73} & 
        \ding{73}\ding{73} & 
        \ding{72}\ding{72} & 
        \ding{73}\ding{73} \\ 
       & POS selectivity& 19.09 & 3.6 & 9.76 & 25.67 & \textbf{43.13} & 18.92 \\ 
         & NER gain & 73.10 & 47.52 & 60.57 & \textbf{77.28} & \textbf{80.18} & 58.46 \\ 
         & NTC accuracy  & 80.25 & 82.47 & 83.28 & \textbf{77.27} & \textbf{94.61} & \textbf{92.76} \\
        \midrule
        \multirow{4}{*}{\textbf{AfriBERTa}}         
        & Language coverage & \ding{72}\ding{72} & 
        \ding{72}\ding{72} & 
        \ding{72}\ding{72} & 
        \ding{73}\ding{73} &
        \ding{73}\ding{73} & 
        \ding{72}\ding{72} \\ 
         & POS selectivity & 18.79 & -0.82 & 14.77 & 18.85 & 30.87 & 14.55 \\ 
         & NER gain & 42.45 & 53.78 & 64.52 & 48.75 & 67.36 & 57.80 \\
           & NTC accuracy & 82.56 & \textbf{84.02} & 87.38 & 74.55 & 81.48 & 92.76 \\ 
         \midrule
        \multirow{4}{*}{\textbf{AfroLM}} 
        & Language coverage & \ding{72}\ding{72} & 
        \ding{72}\ding{72} & 
        \ding{72}\ding{72} &
        \ding{72}\ding{72} & 
        \ding{72}\ding{72} & 
        \ding{72}\ding{72} \\ 
        & POS selectivity  & \textbf{23.64} & 3.37 & \textbf{26.55} & \textbf{27.98} & 38.92 & \textbf{27.28} \\ 
        & NER gain  & 39.45 & 46.33 & 61.21 & 45.90 & 62.07 & 47.97 \\ 
           & NTC accuracy & 76.05 & 76.29 & 85.80 & 69.09 & 80.47 & 92.76 \\
         \bottomrule
    \end{tabular}
    \caption{Best-layer performance for each probing task, with best task performance overall indicated in \textbf{boldface}. We show this alongside model language coverage to indicate how language inclusion improves probe performance. \ding{73}\ding{73} indicates no language included in pretraining or adaptation, \ding{72}\ding{73} shows the language was included in the base model but not in the adapted model, while \ding{72}\ding{72} shows the model was either pretrained or adapted for the language.}
    \label{tab:aggregated_results}
\end{table*}

\section{Conclusion}

This paper presents a systematic analysis of the linguistic knowledge encoded in PLMs for African languages. Through extensive probing experiments across seven PLMs and six typologically diverse African languages, we highlight trends in how PLMs represent syntactic and semantic information.
To contextualise our results we design a control task for POS tagging and employ randomly initialised baselines to compare against NER and NTC probing results. We show that multilingual adaptation reliably enhances hidden representations for target languages. While token-level linguistic knowledge is primarily encoded in middle and deeper layers, sentence-level semantic information is distributed across layers. We find evidence that cross-lingual learning enhances representations for low-resource languages, such as Luganda, but cannot be relied on to do so for under-represented languages, such as Igbo. We hope this work inspires further research at the intersection of interpretability and NLP for African language.

\section*{Limitations}

As discussed in Section \ref{sec:background}, designing control tasks for NER proved challenging. While control tasks are primarily designed for word-level tasks, NER presents complications because named entities often span multiple words. This makes it difficult to apply the typical control task framework in a meaningful way. Instead, we relied on random baselines, which, although commonly used, are known to have certain limitations \citep{10.1162/coli_a_00422, hewitt2019designing}. %Developing appropriate control tasks for NER remains an open problem and a promising area for future work.

In this study, we used the first subword as input for the classifier to align tokens with their hidden representations. However, even the choice of subword pooling strategy can make a difference in probing performance \citep{Acs:2020}. Other pooling strategies, such as last subword, mean pooling, or attention over subwords, could provide different insights, especially for morphologically rich languages with high subword tokenization rates. Future work should systematically compare the effects of different subword pooling strategies across various syntactic and semantic tasks for African languages.
% Time constraints limited the exploration of subword pooling strategies in our probing classifiers, 

\section*{Acknowledgements}

Wisdom Aduah was a Google DeepMind scholar at AIMS South Africa, in the AI for Science Masters program when he conducted this research.

% \section*{Ethical considerations}
% This research is conducted with the goal of improving the understanding and interpretability of language models for African languages. The datasets used are publicly available, and all experiments follow ethical guidelines for NLP research, ensuring transparency and reproducibility. No personally identifiable information is processed, and care has been taken to avoid reinforcing biases inherent in language models. Future work will continue to address fairness, inclusivity, and responsible AI development in multilingual settings.
\bibliography{custom}

\clearpage
\onecolumn
\appendix
\section{Data Information}
%\vspace{-1}
%\FloatBarrier
\label{sec:appendix}
\begin{table*}[!h]
    \centering
    \begin{tabular}{ccccc}
    \hline
         & Bantu (\%) & Volta-Niger (\%) & Afro-Asiatic (\%) & Others (\%)\\
         \hline
         XLM-R & 33.3 &  0.0 & 6.3 & 60.4 \\
         AfroXLMR & 28.0 & 5.8 & 7.4 & 58.8 \\
         AfriBERTa & 0.0 & 7.4 & 16.0 & 76.6 \\
         AfroLM & 32.8 & 9.7 & 18.4 & 39.1\\
         \hline
    \end{tabular}
    \caption{Distribution of African datasets by language family}
    \label{tab:language_distribution_family}
\end{table*}

% \newpage
% \clearpage
\section{Figures}
\label{sec:supplementary_results_figures}
\begin{figure*}[!h]
    \centering
    \makebox[\textwidth][c]{%
        \includegraphics[width=\textwidth]{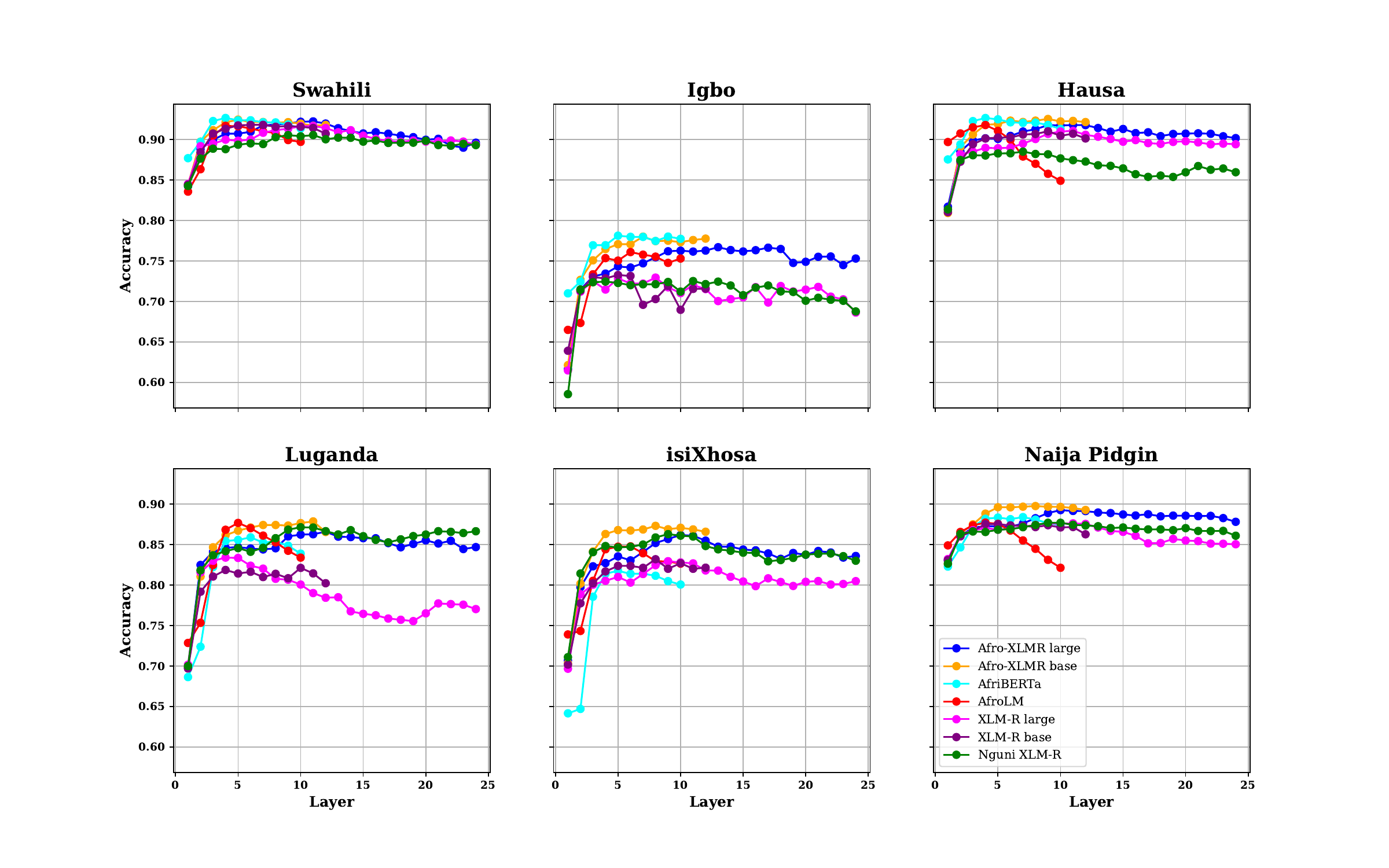}
    }
    \caption{Raw accuracies for POS tagging.}
    \label{fig:raw_pos_accuracies}
\end{figure*}
\begin{figure*}[!t]
    \centering
    \makebox[\textwidth][c]{%
        \includegraphics[width=1\textwidth]{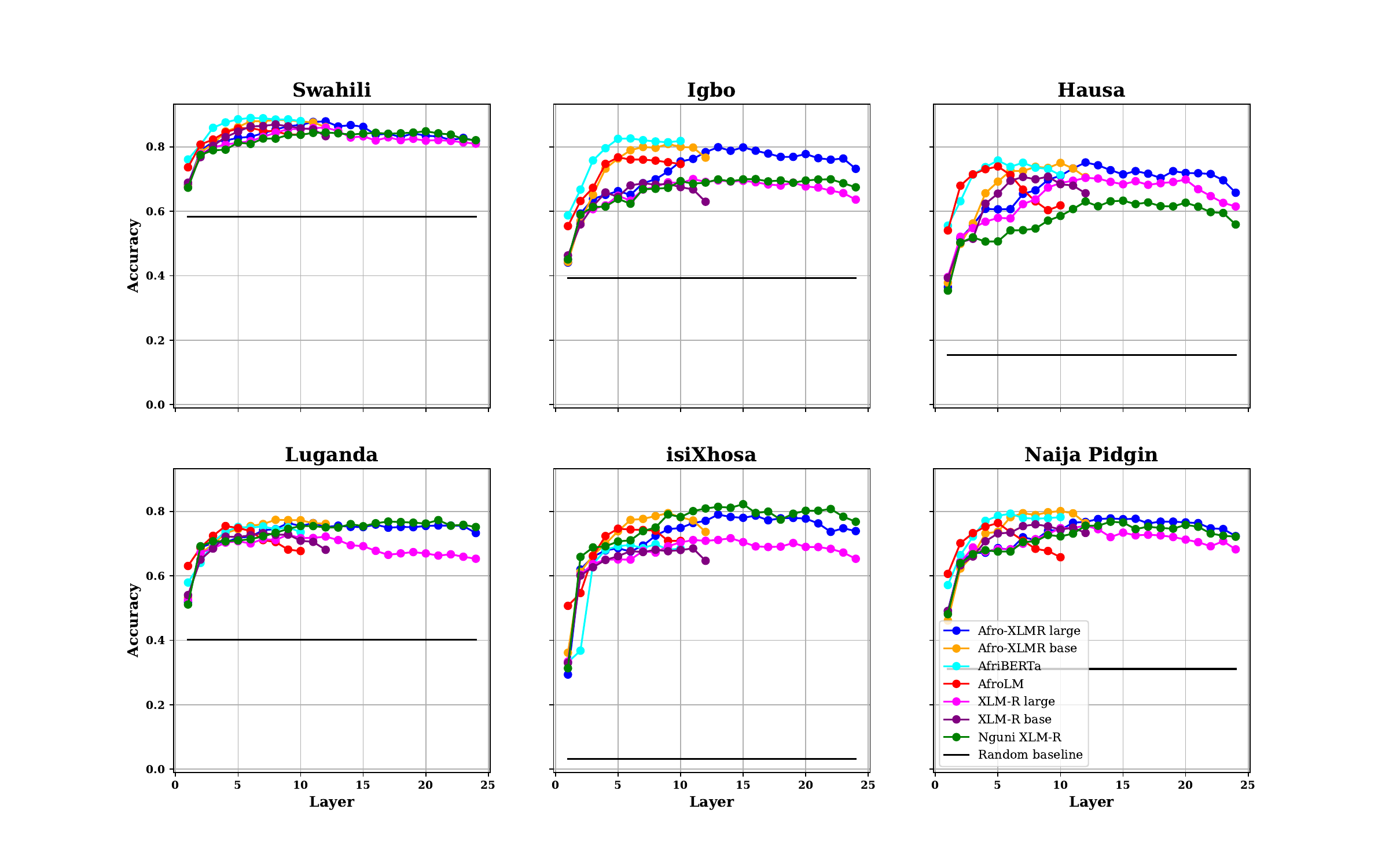}
    }
    \caption{Raw F1-scores for Named Entity Recognition.}
    \label{fig:raw_ner_f1_scores}
\end{figure*}
%\FloatBarrier

%\FloatBarrier
% Bibliography entries for the entire Anthology, followed by custom entries
%\bibliography{anthology,custom}
% Custom bibliography entries only

\end{document}